\documentclass[conference]{IEEEtran}

\IEEEoverridecommandlockouts

\usepackage{cite}
\usepackage{amsmath}
\usepackage{amssymb}
\usepackage{graphicx}
\usepackage{float}
\usepackage{booktabs}
\usepackage{algorithm}
\usepackage{algpseudocode}
\usepackage{xcolor}
\usepackage{textcomp}
\usepackage{listings}
\usepackage{etoolbox}
\usepackage{adjustbox}
\usepackage[hidelinks]{hyperref}
\usepackage{cleveref}
\usepackage{xcolor} 
\usepackage{soul}   
\usepackage{ifsym}
\usepackage{marvosym}

\setkeys{Gin}{width=\columnwidth,keepaspectratio}

\AtBeginEnvironment{table}{\footnotesize}
\lstdefinestyle{ieeecol}{
  basicstyle=\ttfamily\footnotesize,
  columns=fullflexible,
  keepspaces=true,
  breaklines=true,
  breakatwhitespace=false,
  upquote=true,
  frame=single,
  framerule=0.2pt,
  rulecolor=\color{black!25},
  xleftmargin=0.5em,
  xrightmargin=0.0em,
  aboveskip=0.6\baselineskip,
  belowskip=0.6\baselineskip,
  tabsize=2,
  showstringspaces=false,
  captionpos=b
}
\crefname{paragraph}{paragraph}{paragraphs}
\Crefname{paragraph}{Paragraph}{Paragraphs}

\def\BibTeX{{\rm B\kern-.05em{\sc i\kern-.025em b}\kern-.08em
    T\kern-.1667em\lower.7ex\hbox{E}\kern-.125emX}}

\begin{document}

\title{\LARGE \bf
TacEvo: Self-Evolving Architecture Discovery for Robotic Tactile Perception via LLM-Driven Quality-Diversity Search}
\author{ Mohammed AbuSadeh, 
Lan Wei\textsuperscript{\Letter}, 
Dandan Zhang\textsuperscript{\Letter}
\thanks{
Corresponding: l.wei24@imperial.ac.uk and d.zhang17@imperial.ac.uk.

All authors are with the Department of Bioengineering, Imperial-X Initiative, Imperial College London, London, United Kingdom. Code is available at: https://github.com/LannWei/TacEvo.
}
}


\maketitle

\begin{abstract}
Vision-based tactile sensing converts contact-induced surface deformation into images, enabling robots to infer contact forces and fine surface textures that are not accessible through conventional vision alone. However, tactile images are sensor- and physics-specific, so effective architectures often require expert intuition and extensive manual iteration. Existing neural architecture search (NAS) pipelines can reduce this burden, but they are often computationally expensive and restricted to hand-designed search spaces, which limits architectural novelty and diversity.  We introduce TacEvo, a self-evolving architecture discovery framework that improves network designs from downstream feedback. TacEvo uses an LLM to generate code-level mutations and crossovers, and a MAP-Elites quality-diversity loop that preserves diverse elite architectures while preferentially reusing prompts that consistently yield improvements. Exploration is guided by two behavioural descriptors, \textit{Architectural Diversity} and \textit{Efficiency Ratio}, which encourage coverage across structural variations and compute–size trade-offs.
On ViTacTip force regression and grating classification, TacEvo achieves high autonomous generation reliability (96.0\%/94.5\% trainable) and improves best validation fitness over 20 generations by 56.1\%/96.1\%. 
In a 20-seed post-search high-fidelity evaluation, TacEvo matches the expert baseline on force prediction and outperforms it on fine-grained grating classification.
These results suggest that LLM-driven self-evolving search constitutes a practical paradigm for AI-assisted scientific discovery in specialised robotic sensing.

\end{abstract}


\section{Introduction}
\label{sec:introduction}

Robots that manipulate and interact with the physical world must perceive properties that vision alone cannot reliably recover, such as contact location, shear and normal forces, incipient slip, compliance, and fine surface texture~\cite{navarro2023visuo}. Vision-based tactile sensors (VBTSs) bridge this gap by providing high-resolution, dense, and spatially distributed measurements of contact interactions. These sensors capture subtle deformations of a sensing surface, enabling estimation of forces, slips, and textures that are otherwise imperceptible through vision alone~\cite{yuan2017gelsight}. 
The images produced by VBTSs differ fundamentally from natural-image statistics: the most informative cues originate from subtle, geometry-driven changes in deformation patterns and marker displacements~\cite{cong2025taceva}. As a result, achieving high-performance tactile perception typically requires task- and sensor-specific neural architectures~\cite{higuera2024sparsh}. Designing these architectures remains heavily dependent on expert intuition, iterative tuning, and substantial engineering effort.


Neural Architecture Search (NAS) was introduced to automate architecture design by exploring candidate networks and selecting models that optimise task performance under compute and memory constraints~\cite{ren2021comprehensive}. 
 Classical reinforcement learning– and evolution-based NAS approaches have achieved strong performance on mainstream vision benchmarks~\cite{guo2026review}. However, these methods typically require training and evaluating a large number of candidate models, resulting in substantial computational cost. This requirement limits their practicality in many robotics settings, where compute resources are constrained and rapid iteration is often necessary.
 While later work has improved efficiency via weight-sharing and differentiable search, many of these methods still rely on a fixed, hand-specified search space and can be biased toward conventional architectural motifs~\cite{xie2021weight}. 
In specialised sensing domains, such as tactile perception, this emphasis on predefined search operators can restrict novelty and may miss architectures whose inductive biases better match the underlying physics of contact and deformation.


Recent advances in large language models (LLMs) suggest an alternative paradigm for neural architecture design. Instead of searching within a predefined set of operations, an LLM can directly generate architectural variations at the code level, enabling more flexible and open-ended exploration of the design space~\cite{wang2023review}.
This opens the search space substantially and leverages prior architectural knowledge implicitly encoded in large-scale code and text corpora~\cite{dong2025survey}. 
However, open-ended generation also introduces practical challenges: 
(i) the search can collapse to narrow families of architectures without explicit diversity pressure~\cite{chen2021evaluating}, 
(ii) generated code may be syntactically correct yet untrainable or incompatible with the task’s I/O constraints~\cite{romera2024mathematical}, 
and (iii) naive exploration can waste evaluations by repeatedly proposing similar candidates~\cite{ying2019bench}. 
To be useful for robotics, LLM-driven architecture discovery must therefore be grounded in downstream task feedback, reliable as a code-producing operator, and explicitly designed to maintain a broad set of viable solutions rather than converging to a single design.

In this work, we cast tactile architecture design as a self-evolving discovery process: the system repeatedly proposes architectural hypotheses, tests them via downstream-task training, and updates an explicit memory of what works both in the architectures it retains and in the generation strategies that produced them. 
We introduce TacEvo, an LLM-driven architecture evolutionary framework coupled with quality-diversity (QD) optimisation for vision-based tactile sensing. 
TacEvo treats the LLM as a code-level mutation/crossover operator that edits the network, while a Centroidal Voronoi Tessellation MAP-Elites (CVT MAP-Elites)~\cite{cvtmapelites} archive performs selection and long-term retention of diverse, high-quality candidates. 
Crucially, TacEvo maintains two cooperative archives: a network archive of elite architectures and a prompt archive that ranks prompting strategies; prompts receive a curiosity signal when they generate architectures accepted into the network archive, enabling the system to co-evolve its generator policy alongside the architectures it discovers. 
To make QD effective in this open-ended, code-level generation setting, we introduce two behavioural descriptors: Architectural Diversity and Efficiency Ratio. These descriptors guide the archive toward illuminating solutions across structural heterogeneity and compute-size trade-offs, preventing the search from collapsing onto a single architectural motif.


We evaluate TacEvo using the ViTacTip~\cite{fan2024vitactip}, a VBTS, on two representative tasks spanning regression and fine-grained classification: 3-axis force prediction and 7-class grating classification \cite{zhang2025design}. 
TacEvo runs for 20 generations with 50 LLM-proposed candidates per generation, using low-fidelity training (10 epochs) to provide rapid feedback during search, followed by a post-search high-fidelity evaluation that fully trains the top discovered models across multiple random seeds for comparison against a strong expert-designed baseline. 
Empirically, TacEvo exhibits key properties of a self-evolving system: it consistently generates trainable networks across generations, progressively improves the best validation fitness, and maintains broad archive growth and coverage, indicating sustained exploration of diverse architectural niches.
Under the high-fidelity evaluation protocol, 
the discovered architectures are competitive with the expert baseline on force prediction, with the best TacEvo variants statistically indistinguishable from the baseline, and significantly outperform the baseline on grating classification.


This paper makes the following contributions:
\begin{enumerate}
    \item A self-evolving, LLM-in-the-loop architecture searching system for tactile perception, closing the hypothesis–evaluation–memory loop by using downstream task performance feedback to iteratively refine architecture proposals.
    \item A dual-archive QD formulation with a network archive and a prompt archive, enabling co-evolution of architectures and generation strategies for reliable and diverse search.
    \item Two architecture-centric behavioural descriptors: Architectural Diversity and Efficiency Ratio that enable CVT MAP-Elites to illuminate tactile architectures across structural and efficiency trade-offs, rather than collapsing to a single motif.
    \item An empirical study on two ViTacTip tactile tasks demonstrating stable autonomous generation, quality-diversity exploration, and competitive-to-superior downstream performance relative to the expert baseline without manual architecture tuning.
\end{enumerate}

\section{Related Work}
\label{sec:related_work}
\subsection{Closed-loop Architecture Discovery}
Many scientific and engineering advances follow a closed loop: propose a hypothesis, test it, and update the next hypothesis based on measured feedback. 
In machine learning, AutoML~\cite{he2021automl,barbudo2023eight} and NAS~\cite{elsken2019neural,zhou2025design} instantiate this loop by generating candidate models, evaluating them, and iteratively refining the search toward better architectures. 
In robotics, self-adaptive systems further emphasise continual improvement under changing conditions, where maintaining and reusing a repertoire of solutions can enable rapid adaptation without restarting optimisation from scratch~\cite{alberts2025software}. 
Despite this progress, most architecture-discovery pipelines still focus on single-objective optimisation. They also offer limited support for preserving multiple viable designs, even though such diversity is important when deployment constraints, sensing conditions, and task requirements vary across robots and environments.

Related work has also studied self-evolution at the sensing-system level. 
For example, Xu et al.~\cite{xu2022optimization} optimised forcemyography sensor placement for arm movement recognition using greedy local search, while Zenatti et al.~\cite{zenatti2016optimal} optimised passive sensor placement for robot localisation. 
These works show that heuristic search can improve complete sensing systems when the sensor configuration and objective are explicitly defined. 
In contrast, TacEvo keeps the tactile sensor hardware fixed and evolves the neural architecture that interprets tactile images,
targeting self-improvement at the perception-model level.

\subsection{LLM-based Model Generation}
Transformer-based LLMs have demonstrated strong code-generation capabilities, motivating their use as open-ended proposal engines that can write executable network definitions rather than selecting from a fixed operator set~\cite{wang2025ai}. 
Recent work reframes the LLM as an evolutionary operator: given a parent architecture and an instruction, the model produces a code-level mutation, or merges two parents via crossover~\cite{evoprompting}. 
Building on this view, LLMatic couples LLM proposals with evolutionary search to discover competitive architectures with relatively few evaluations~\cite{LLMatic}. 
However, unconstrained generation can yield invalid or untrainable programs, and search can collapse to repetitive patterns without explicit diversity pressure. 
These limitations highlight the need for systematic validation, feedback-driven guidance, and mechanisms that stabilise and diversify generation when LLMs are used as operators inside autonomous discovery loops.

\subsection{Quality-Diversity (QD) Algorithm for Exploration}
QD methods, exemplified by MAP-Elites, aim to illuminate a search space by filling an archive with the best-performing solution in each behavioural niche, producing a diverse repertoire of elites rather than a single optimum~\cite{mapelites}. 
CVT MAP-Elites scales this concept by using centroidal Voronoi tessellations to create well-distributed niches in higher-dimensional descriptor spaces~\cite{vassiliades2017using}. 
In robotics design, QD repertoires have enabled fast adaptation by offering diverse behaviours that can be selected and refined~\cite{cully2015robots}. 
In model and architecture discovery, QD provides a principled way to retain multiple high-quality architectural motifs while exploring trade-offs instead of converging prematurely to one design family~\cite{LLMatic}. 
A key practical challenge is descriptor design: behavioural descriptors must meaningfully capture the diversity dimensions that matter for the domain, otherwise the archive may fail to represent genuinely distinct solutions and the search degenerates into near-single-objective optimisation.

\begin{figure*}[!t]
    \centering
    \includegraphics[width=1\linewidth]{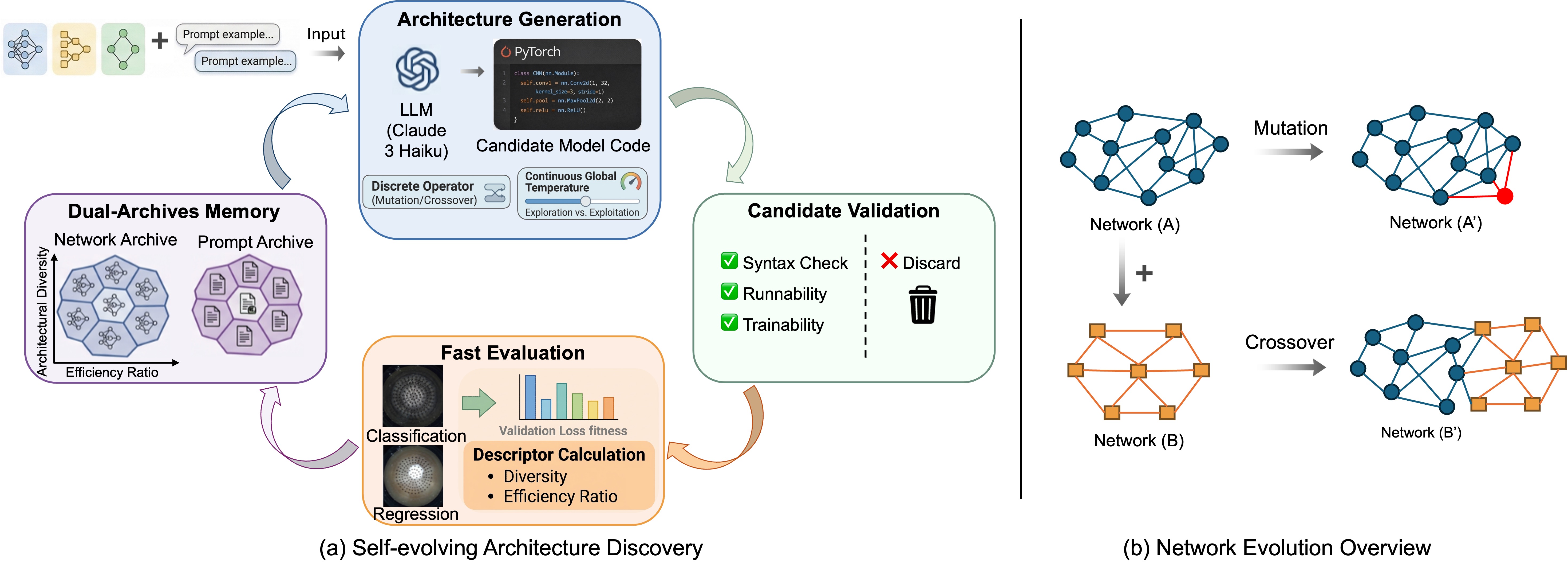}
    \vspace{-0.7cm}
    \caption{Overview of TacEvo. (a) Closed-loop self-evolving architecture discovery with LLM-based code generation, candidate validation, low-fidelity evaluation, and dual-archive memory. (b) LLM-driven mutation and crossover for network evolution.}
    \label{fig:overvire}
    \vspace{-0.5cm}
\end{figure*}

Building on these threads, we propose an LLM-driven model evolutionary framework coupled with CVT MAP-Elites to realise closed-loop, self-evolving architecture discovery for vision-based tactile perception. 
While generic coding agents such as AlphaEvolve~\cite{novikov2025alphaevolve} provide a broader framework for evolving arbitrary executable programs, TacEvo focuses on tactile-specific architecture discovery under fixed sensor input/output constraints, downstream tactile-task evaluation, and QD descriptors designed to capture structural and efficiency trade-offs in tactile perception models.

\section{Methodology}
\label{sec:method}

\subsection{Problem Setup and Objective}
\label{subsec:problem}

We consider a vision-based tactile learning problem with a training split $D_{\mathrm{train}}$ and a validation split $D_{\mathrm{val}}$. 
Let $a\in\mathcal{A}$ denote a candidate neural architecture, and let $w_a^*$ be the weights obtained by optimising $a$ on $D_{\mathrm{train}}$. 
The architecture discovery objective is to find an architecture that minimises the downstream validation loss:
\begin{equation}
 a^* \;=\; \arg\min_{a\in\mathcal{A}} \; \mathcal{L}_{\mathrm{val}}\!\left(a;\,w_a^*,\,D_{\mathrm{val}}\right).
\label{eq:objective}
\end{equation}

TacEvo realises this objective as a closed loop that repeatedly proposes code-level architectural hypotheses, tests them via training on the downstream task, and updates an explicit memory of what works. 
This memory consists of a repertoire of diverse high-performing architectures, together with bookkeeping signals about which prompting configurations tend to yield archive-improving candidates.

\subsection{TacEvo Architecture Discovery}
\label{subsec:closedloop}

\subsubsection{System Overview}
\label{subsubsec:overview}

TacEvo couples an LLM-based program generator with quality-diversity optimisation to implement a closed-loop architecture discovery process, as shown in Algorithm~\ref{alg:tacevo}. 
At each generation, the pipeline samples elite parent architectures from a network archive, samples a single evolutionary operator for the generation (mutation or crossover), invokes the LLM to produce executable code for candidate child architectures, validates the generated code, and evaluates valid candidates using a low-fidelity training budget to obtain fast downstream feedback. 
Candidates compete for niches in a CVT MAP-Elites archive defined over behaviour descriptors; archive insertion and replacement provide selection pressure while archive-wide diversity is maintained by construction. 
In addition, the implementation maintains a prompt archive that records prompt configurations and their outcomes; however, mutation prompts are drawn from a per-generation randomly sampled prompt pool rather than being selected from the prompt archive.

\subsubsection{Search Space and Baseline Constraints}
\label{subsubsec:searchspace}

TacEvo performs search at the level of executable network code, but under strict task compatibility constraints. 
The search is seeded by a manually designed baseline CNN $a_0$ (the same input resolution and channel convention as the tactile images, and a task-specific output layer), which serves as the genetic foundation for evolution. 
All candidates must preserve the I/O contract of the downstream task (e.g., a fixed-dimensional regression output or a fixed number of class logits), and must implement a valid forward pass with consistent tensor shapes. 
Within these constraints, candidates may alter depth, width, connectivity, pooling and normalisation choices, activation functions, and higher-level motifs such as residual/skip connections or attention blocks, provided the resulting program remains trainable under standard gradient-based optimisation.

\begin{algorithm}[!t]
\caption{TacEvo Self-evolving Architecture Discovery}
\label{alg:tacevo}
\begin{algorithmic}[1]
\State Initialise Network Archive $\mathcal{A}_N$ from mutations of baseline $a_0$; initialise Prompt Archive $\mathcal{A}_P$; set global temperature $T_1$.
\For{$g=1,\dots,G$}
  \State Sample a single operator $\mathrm{op}_g\in\{\mathrm{mut},\mathrm{cross}\}$ for the generation.
  \State Sample a prompt pool $\mathcal{P}_g$ uniformly at random from a predefined prompt library.
  \State Generate a batch of $B$ candidate architectures $\mathcal{B}_g$ using $\mathrm{op}_g$ and temperature $T_g$.
  \State Validate $\mathcal{B}_g$ for syntax, compilability, and trainability to obtain the valid subset $\tilde{\mathcal{B}}_g$ (invalid candidates are discarded; no resampling).
  \ForAll{$a'\in\tilde{\mathcal{B}}_g$}
    \State Evaluate fitness $f(a')$ via low-fidelity training; compute descriptors $d(a')$; map to niche $n$.
    \State Insert/replace $\mathcal{A}_N[n]$ if $a'$ improves fitness.
    \If{$\mathrm{op}_g=\mathrm{mut}$}
      \State Let $s=(i,T_g)$ be the prompt configuration used to generate $a'$ (prompt index $i$ sampled from $\mathcal{P}_g$).
      \State Update $\mathcal{A}_P$ using the binary success signal $r(s)$ and record curiosity $c(s)$.
    \EndIf
  \EndFor
  \State Update $T_{g+1} \leftarrow \mathrm{TempUpdate}(T_g, \mathcal{A}_N)$.
\EndFor
\State \textbf{return} $\mathcal{A}_N,\mathcal{A}_P$.
\end{algorithmic}
\end{algorithm}

\subsubsection{LLM as an Evolutionary Operator}
\label{subsubsec:llm_operator}
TacEvo treats the LLM as a stochastic evolutionary operator that edits code. 
Generation is driven by a task-specific system prompt that enforces formatting and I/O requirements, and an instruction prompt that specifies the intended architectural transformation. 
The evolutionary operator is sampled \emph{once per generation}: with probability $p_m$ the generation applies mutation, and with probability $1-p_m$ it applies crossover. 
Under mutation, each candidate is produced by selecting an elite parent architecture $a$ from the network archive and sampling an instruction prompt $p$ from the generation's prompt pool $\mathcal{P}_g$; the LLM outputs a mutated child $a'$. 
Under crossover, each candidate is produced by selecting two elite parents $(a_1,a_2)$ from the network archive and using a fixed crossover instruction; the LLM outputs a hybrid child $a'$. 
Because the code generation may fail in subtle ways, every LLM output passes a validation gate before evaluation; a candidate is considered valid only if it is syntactically correct, instantiable, and trainable (i.e., it supports forward/backward computation under the task loss). 
Invalid candidates are discarded prior to training.

\subsubsection{Quality-Diversity Optimization}
\label{subsubsec:qd}

TacEvo uses CVT MAP-Elites to turn downstream feedback into a persistent repertoire of solutions. 
Each evaluated architecture $a'$ is assigned a scalar fitness $f(a')$ (the final validation loss from low-fidelity training) and a behavioural descriptor vector $d(a')$ (Section~\ref{subsubsec:descriptors}). 
CVT partitions the descriptor space into niches using a fixed set of centroids, and the network archive $\mathcal{A}_N$ stores at most one elite per niche. 
A candidate $a'$ is inserted into niche $n$ if the niche is empty or if $f(a')$ is better than the current occupant. 
This archive-based competition prevents collapse to a single architectural motif and explicitly preserves diverse elites across structural and efficiency regimes.

\subsubsection{Network Archive Descriptors}
\label{subsubsec:descriptors}

To make QD search informative for architecture discovery in tactile perception, we define two descriptors and min-max normalise them to $[0,1]$ before CVT mapping. 
The first descriptor, \emph{Architectural Diversity}, summarises structural heterogeneity by combining operation-type entropy, connection density, and layer heterogeneity. 
Let $p_i$ be the fraction of occurrences of operation type $i$ among $N$ operation types (excluding trivial containers and simple activations). 
The normalised operation entropy is
\begin{equation}
 H_{\mathrm{norm}} \;=\; \frac{-\sum_{i=1}^{N} p_i \log_2 p_i}{\log_2 N}.
\label{eq:entropy}
\end{equation}
Connection density measures how richly layers are connected (including skip connections):
\begin{equation}
 \mathrm{Density} \;=\; \frac{C_{\mathrm{actual}}}{C_{\mathrm{possible}}}.
\label{eq:density}
\end{equation}
Layer heterogeneity is computed as the mean coefficient of variation across $P$ layer properties $x_k$ (e.g., channels, kernel sizes, hidden widths):
\begin{equation}
 \mathrm{CV} \;=\; \frac{1}{P}\sum_{k=1}^{P}\frac{\sqrt{\mathrm{Var}(x_k)}}{\mathrm{Mean}(x_k)}.
\label{eq:cv}
\end{equation}
The Architectural Diversity score is then
\begin{equation}
 \mathrm{ArchDiv} \;=\; \frac{H_{\mathrm{norm}} + \mathrm{Density} + \mathrm{CV}}{3}.
\label{eq:archdiv}
\end{equation}
The second descriptor, \emph{Efficiency Ratio}, captures a compute--structure trade-off by relating FLOPs to the combined structural cost of parameter count and width-to-depth ratio:
\begin{equation}
 \mathrm{EffRatio} \;=\; \frac{\mathrm{FLOPs}}{\mathrm{Params}\times (\mathrm{W/D})}.
\label{eq:effratio}
\end{equation}
Here $\mathrm{W/D}$ is computed as the maximum layer width divided by the network depth. Higher $\mathrm{EffRatio}$ corresponds to architectures that are more computationally dense relative to size and shape cost, enabling the archive to illuminate diverse efficiency regimes rather than optimising only accuracy.

\subsection{Prompt Tracking, Exploration Control, and Fast Feedback}
\label{subsec:policy_feedback}

\subsubsection{Prompt Archive and Curiosity Signals}
\label{subsubsec:prompt_policy}

TacEvo maintains a prompt archive $\mathcal{A}_P$ as a parallel CVT MAP-Elites archive over prompt configurations $s=(i,T)$, where $i$ denotes the prompt index and $T$ is the temperature used for generation. 
Prompts used for mutation are sampled from a per-generation prompt pool $\mathcal{P}_g$ drawn uniformly at random from the prompt library; 
$\mathcal{A}_P$ is updated to record which prompt configurations were productive.
After a mutation-generated candidate $a'$ is evaluated and mapped to niche $n$ in the network archive, we define a binary prompt fitness
\begin{equation}
 r(s) \;=\; \mathbb{I}_{\mathrm{ins}}(a')\in\{0,1\},
\label{eq:prompt_fitness}
\end{equation}
where $\mathbb{I}_{\mathrm{ins}}(a')=1$ iff $a'$ is inserted into (or replaces the occupant of) its network archive niche. 
This binary signal is used for niche competition in the prompt archive. 
In addition, we compute an auxiliary curiosity score that assigns a penalty to unproductive prompt configurations and rewards successful ones proportionally to their achieved loss:
\begin{equation}
 c(s) \,=\, (1-r(s))\cdot 0.5 \; + \; r(s)\cdot\left(1 + \frac{1}{1+f(a')}\right).
\label{eq:curiosity}
\end{equation}
This yields $c(s)\in\{0.5\}\cup(1,2]$ and provides a more graded measure of prompt usefulness for analysis and logging.

\subsubsection{Exploration Control via Temperature Scheduling}
\label{subsubsec:temperature}

Temperature scheduling provides a continuous control knob for balancing exploration and refinement. 
TacEvo maintains a global temperature state $T_g\in[0,1]$ updated once per generation from the progress of the network archive. 
Let $f^\star_g$ denote the best fitness (lowest validation loss) observed in $\mathcal{A}_N$ after generation $g$. 
We update temperature using a progress-conditioned rule:
\begin{equation}
 T_{g+1}=\mathrm{clip}\!\left(T_g + \eta\cdot \mathbb{I}[f^\star_g \le f^\star_{g-1}] - \eta\cdot \mathbb{I}[f^\star_g > f^\star_{g-1}],\, 0,\, 1\right),
\label{eq:temp_update}
\end{equation}
where $\eta$ is a step size and $\mathrm{clip}(\cdot)$ enforces the valid range. 
During candidate generation we apply a small intra-generation perturbation $T=\mathrm{clip}(T_g+\epsilon,0,1)$ with $\epsilon$ drawn from a narrow zero-mean distribution to increase proposal diversity.

\subsubsection{Low-Fidelity Evaluation for Fast Feedback}
\label{subsubsec:low_fidelity}

To scale evaluation across generations, TacEvo uses low-fidelity training as a fast experimental proxy for downstream utility. 
Each validated candidate architecture is trained for a short budget of $E$ epochs using a fixed optimiser and task loss, and its fitness is defined as the final validation loss $f(a')=\mathcal{L}_{\mathrm{val}}(a'; D_{\mathrm{val}})$ from this run. 
This fitness drives network archive competition and, through $r(s)$ in Eq.~(\ref{eq:prompt_fitness}), provides a binary success signal for updating the prompt archive; invalid candidates are discarded prior to this evaluation and are not resampled.

\section{Experiments}
\label{sec:experiments}

\subsection{Experiments Setting}
\subsubsection{Sensor, Tasks, and Datasets}
\label{subsec:datasets}

\begin{figure}[!t]
    \centering
    \includegraphics[width=1\linewidth]{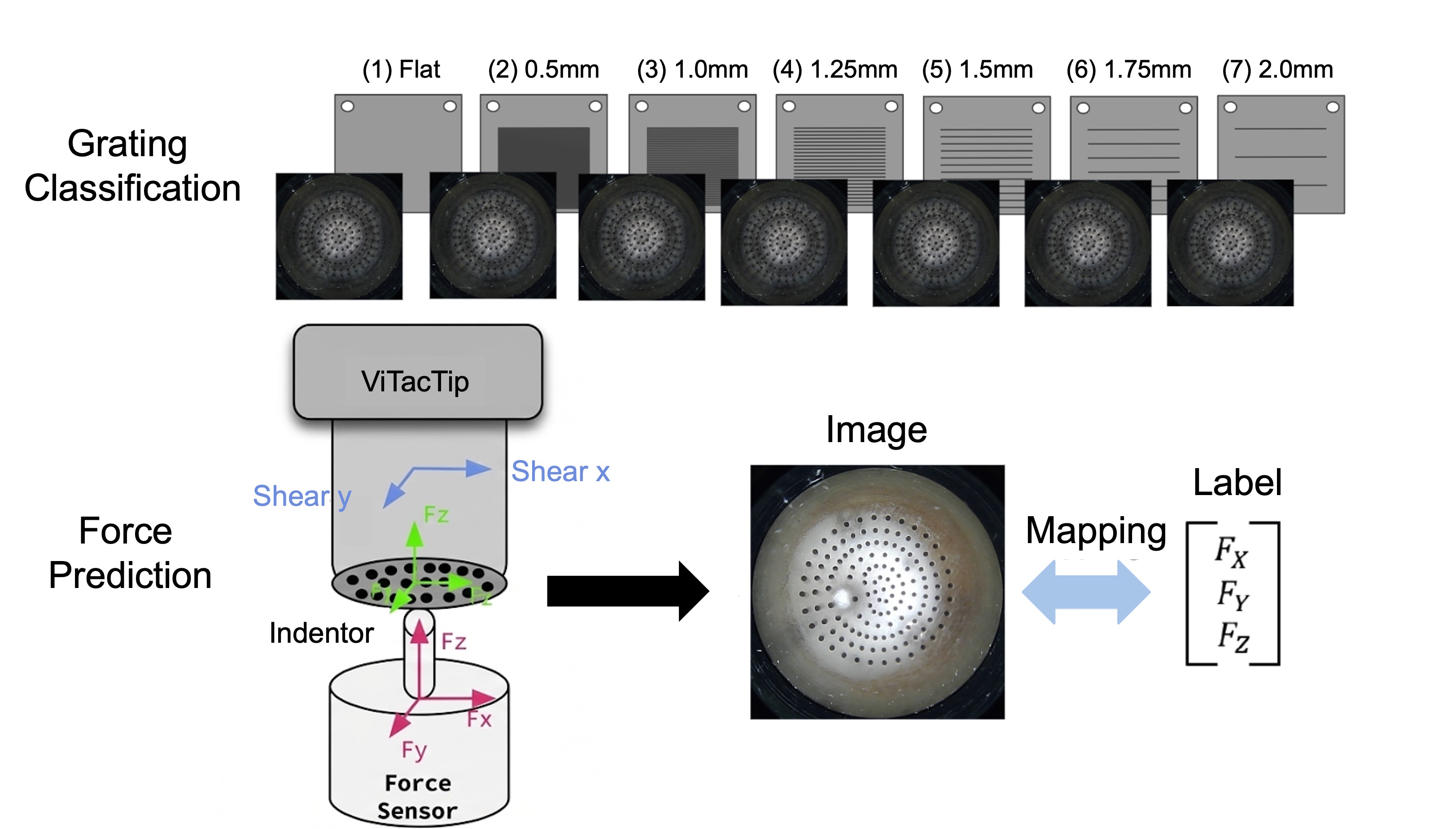}
    \vspace{-0.5cm}
    \caption{Data representation for the two ViTacTip tactile tasks: grating classification and 3-axis force prediction.}
    \label{fig:vitatip_setup}
    \vspace{-0.4cm}
\end{figure}

We evaluate TacEvo on vision-based tactile perception using the ViTacTip sensor.
ViTacTip captures contact-induced deformation of a compliant surface as an image stream, enabling learning-based inference of physical interaction signals.
The experimental setup and representative inputs for the two target tasks are shown in Fig.~\ref{fig:vitatip_setup}. 
We conduct two downstream tasks that span regression and fine-grained classification.
Force Prediction predicts a 3-axis force vector from a tactile image, using a dataset of 3{,}000 labelled images.
Grating Classification distinguishes fine textures across 7 grating-spacing classes using 3{,}507 images evenly distributed across classes.
For both tasks, images are normalised and resized to a uniform $256\times256$ resolution and split into 80\%/10\%/10\% train/validation/test partitions to ensure consistent evaluation across architectures.

\subsubsection{Baseline and Training Recipes}
\label{subsec:baseline_train}

TacEvo is seeded with a CNN baseline, which consists of three convolutional blocks followed by a fully connected head, with the final output layer adjusted to the task (3 outputs for force regression and 7 logits for grating classification).
During the evolutionary search, candidate architectures are evaluated with a low-fidelity budget to produce a fast fitness signal (Section~\ref{subsec:protocol}).

\subsubsection{Search Configuration}
\label{subsec:search_config}

TacEvo is run for $G{=}20$ generations, with Claude 3 Haiku~\cite{anthropic2024claude3} as the LLM proposing up to $B{=}50$ candidate architectures per generation. 
At the beginning of each generation, the pipeline samples an evolutionary operator, mutation with probability 0.85 or crossover with probability 0.15; the chosen operator is then applied to all proposals within that generation. 
For mutation generations, a fresh pool of instruction prompts is sampled uniformly at random from a predefined prompt library, and each candidate mutation selects its prompt from this pool. 
All generated candidates are passed through code validation for syntactic correctness, compilability, and trainability; candidates that fail validation are discarded, and therefore, the number of evaluated networks is strictly smaller than $B$. 
TacEvo uses a CVT MAP-Elites network archive over two behavioural descriptors: Architectural Diversity and Efficiency Ratio (Section~\ref{sec:method}) and a parallel prompt archive that records prompt outcomes.

\subsubsection{Search-phase Evaluation Protocol}
\label{subsec:protocol}

\begin{figure}[!t]
    \centering
    \includegraphics[width=1\linewidth]{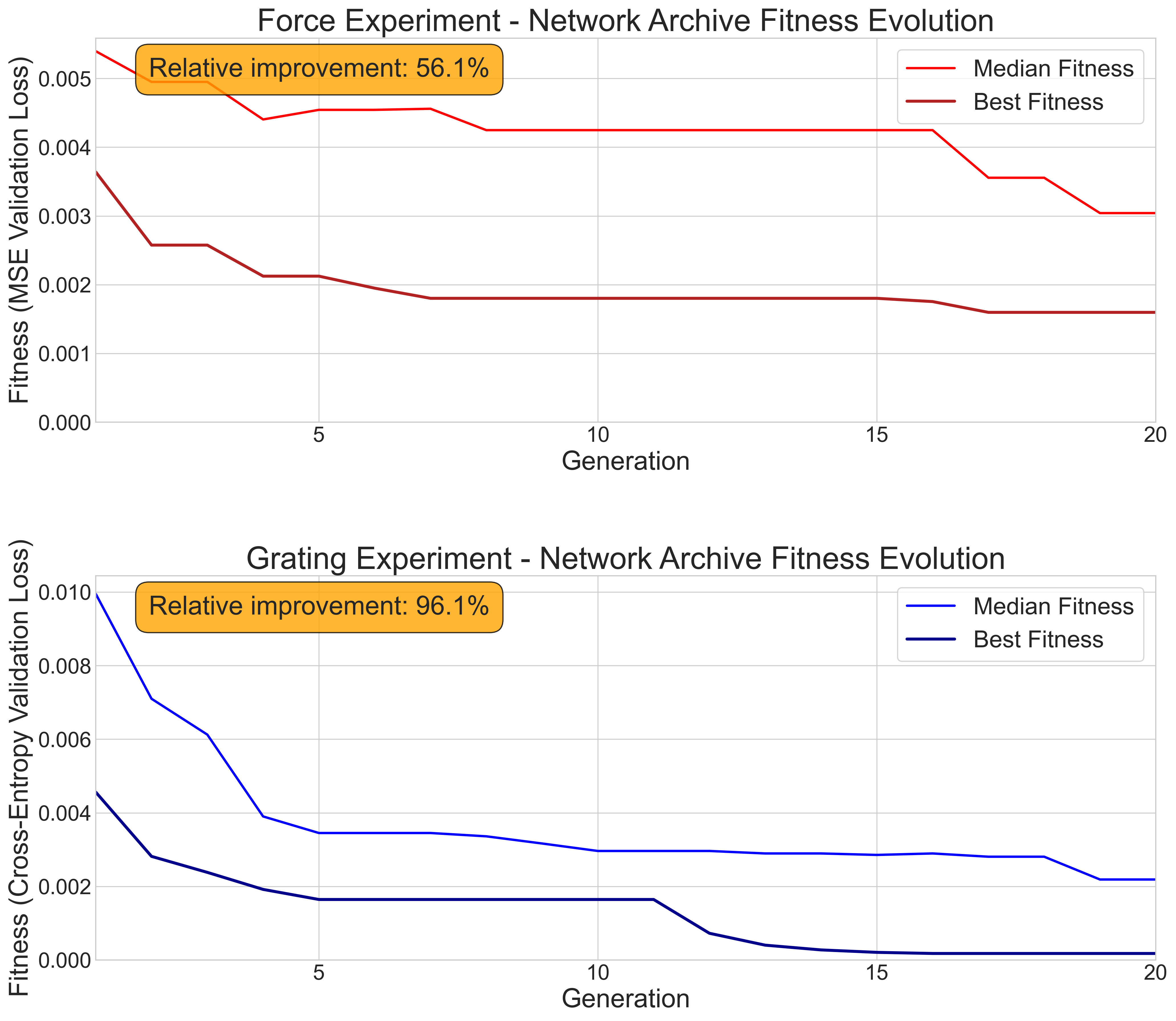}
    \vspace{-0.5cm}
    \caption{Best and median archive fitness over 20 generations under low-fidelity evaluation. TacEvo improves best fitness by 56.1\% on Force Prediction (MSE, top) and 96.1\% on Grating Classification (cross-entropy, bottom).}
    \vspace{-0.5cm}
\label{fig:fitness_evolution}
\end{figure}

We evaluate the search process using search-time signals only. 
Each validated candidate is trained for 10 epochs and assigned a fitness equal to its final validation loss; lower fitness indicates better downstream performance. 
We characterise the search along three axes: (i) quality, tracked by the best and median fitness in the archive over generations;
(ii) reliability, measured by the number of candidates that pass validation and remain trainable per generation; 
and (iii) diversity/illumination, assessed by archive heatmaps and archive growth/coverage over time.

\subsection{Results}
\label{sec:results}
\begin{figure}[!t]
    \centering
    \includegraphics[width=1\linewidth]{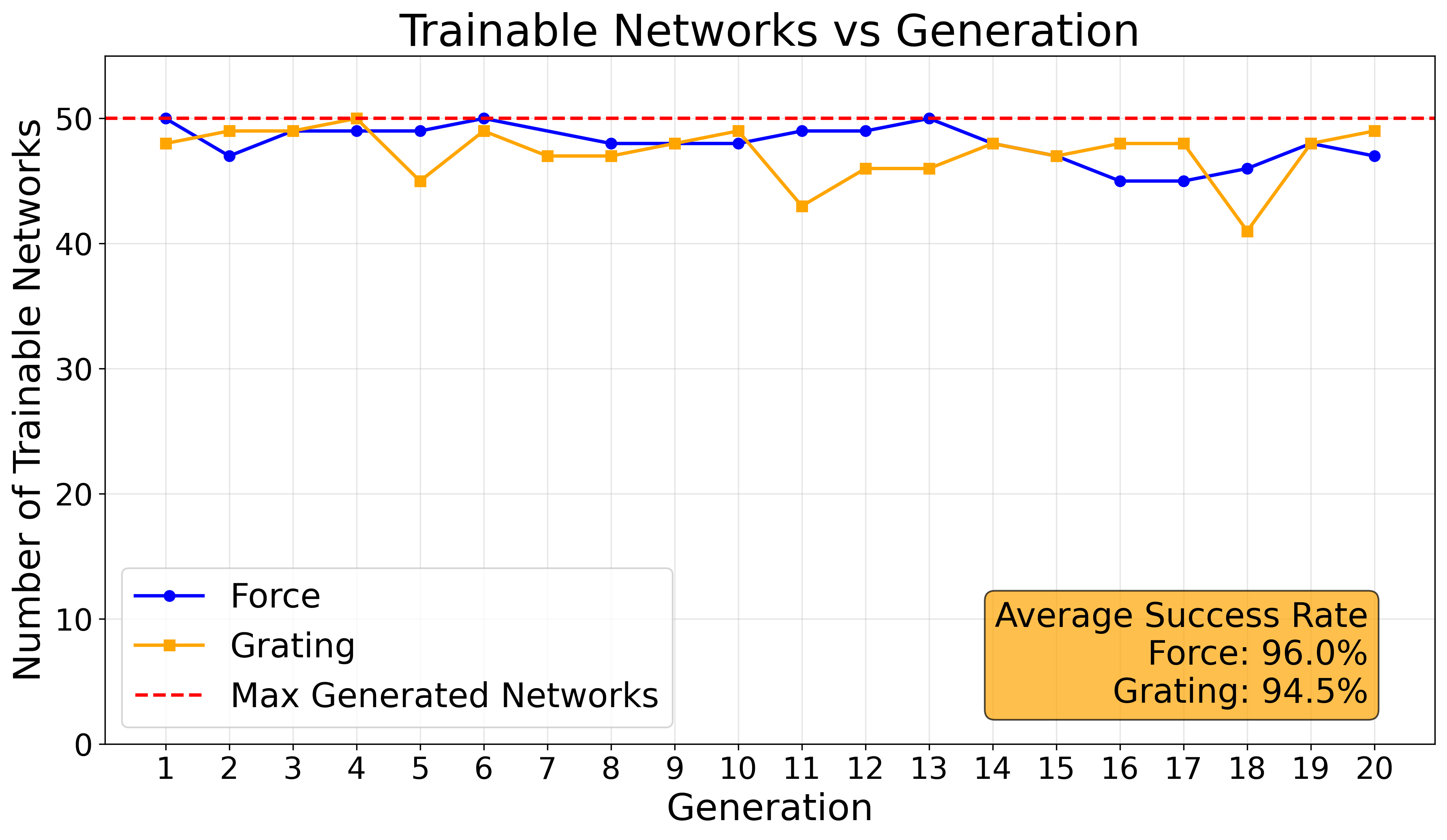}
    \vspace{-0.5cm}
    \caption{Number of valid, trainable networks per generation. Candidates must pass syntax/compilation checks and be trainable under the task loss. }
\label{fig:trainable_rate}
\end{figure}

\begin{figure}[!t]
    \centering
    \includegraphics[width=1\linewidth]{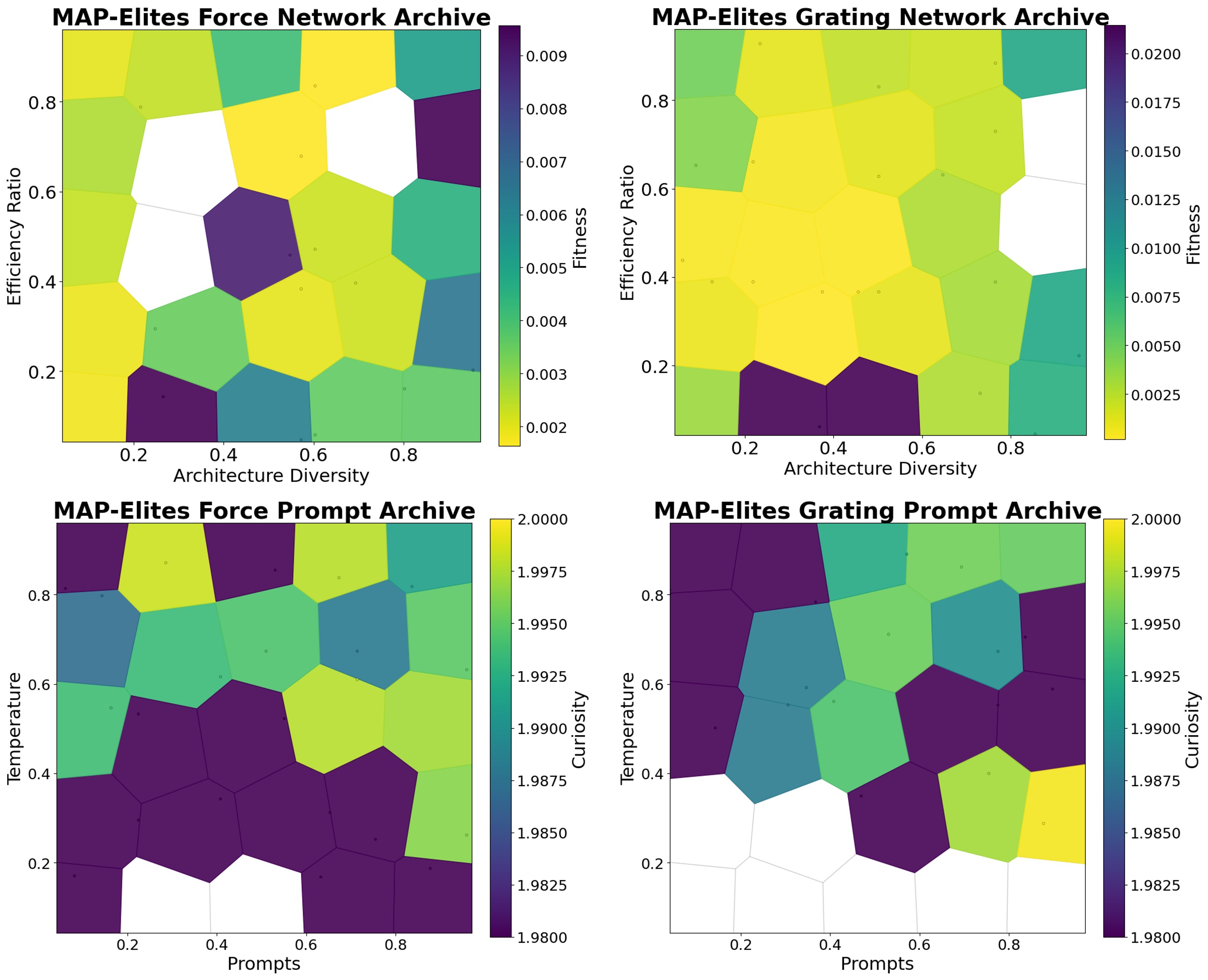}
    \vspace{-0.4cm}
    \caption{Four CVT maps showing the final state of the network and prompt archives for both tasks. The heatmaps visualise the distribution of elite individuals. Brighter cells indicate higher quality (lower fitness for network archives, higher curiosity for prompt archives).}
    \vspace{-0.4cm}
\label{fig:archive_heatmaps}
\end{figure}

\subsubsection{Search-phase Dynamics: Evidence of Self-evolving Discovery}
\label{subsec:search_dynamics}

Fig.~\ref{fig:fitness_evolution} shows the evolution of best and median fitness (validation loss) over 20 generations for both tasks. 
TacEvo exhibits a clear downward trend, indicating that downstream feedback steadily guides the discovery process toward higher-quality architectures. 
The largest improvements occur within the first 7-10 generations, after which the search begins to stabilise, suggesting efficient navigation of the architecture space under the given budget. 
Overall, the relative improvement in the best fitness reaches 56.1\% for Force Prediction and 96.1\% for Grating Classification by the final generation.

Beyond quality, TacEvo maintains high generation reliability. 
As shown in Fig.~\ref{fig:trainable_rate}, the LLM produces a consistently high number of trainable candidates per generation (defined as passing validation and supporting training), with average success rates of 96.0\% for Force and 94.5\% for Grating. 
This indicates that the closed-loop system can operate autonomously at scale without frequent manual intervention for code failures.

\begin{figure}[!t]
    \centering
    \includegraphics[width=1\linewidth]{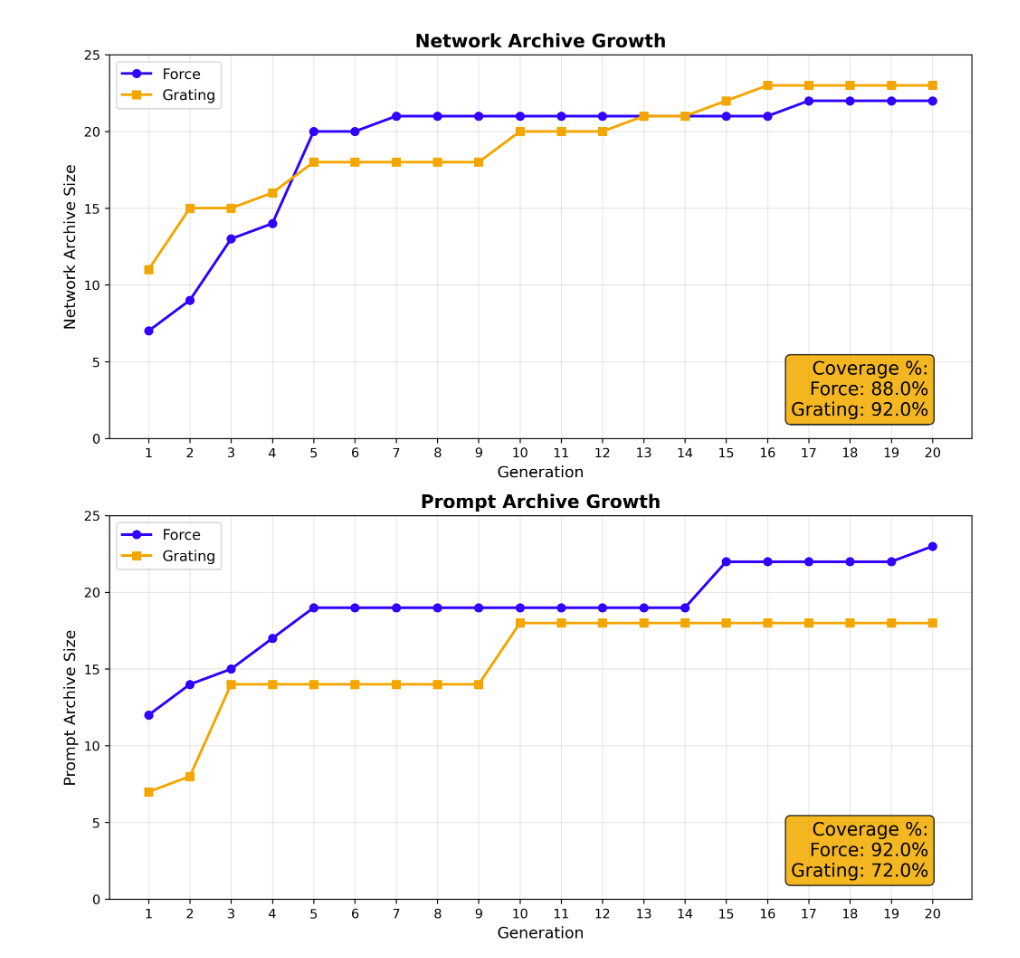}
    \vspace{-0.4cm}
    \caption{Archive growth over 20 generations, measured by the number of filled niches in the network and prompt archives for Force and Grating.}
    \vspace{-0.4cm}
\label{fig:archive_growth}
\end{figure}

Finally, TacEvo demonstrates continuous exploration rather than collapsing to a single architectural motif. 
The final CVT archive heatmaps in Fig.~\ref{fig:archive_heatmaps} show elites distributed broadly across the descriptor space, and Fig.~\ref{fig:archive_growth} quantifies the steady increase in filled niches throughout the run. 
In particular, the final Network Archive coverage reaches 88.0\% (Force) and 92.0\% (Grating), confirming that the QD mechanism successfully illuminates diverse regions of the architecture space while continuing to improve quality.

\subsubsection{Post-search High-fidelity Evaluation: Force Prediction}
\label{subsec:force_results}

\begin{figure}[!t]
    \centering
\includegraphics[width=1\linewidth]{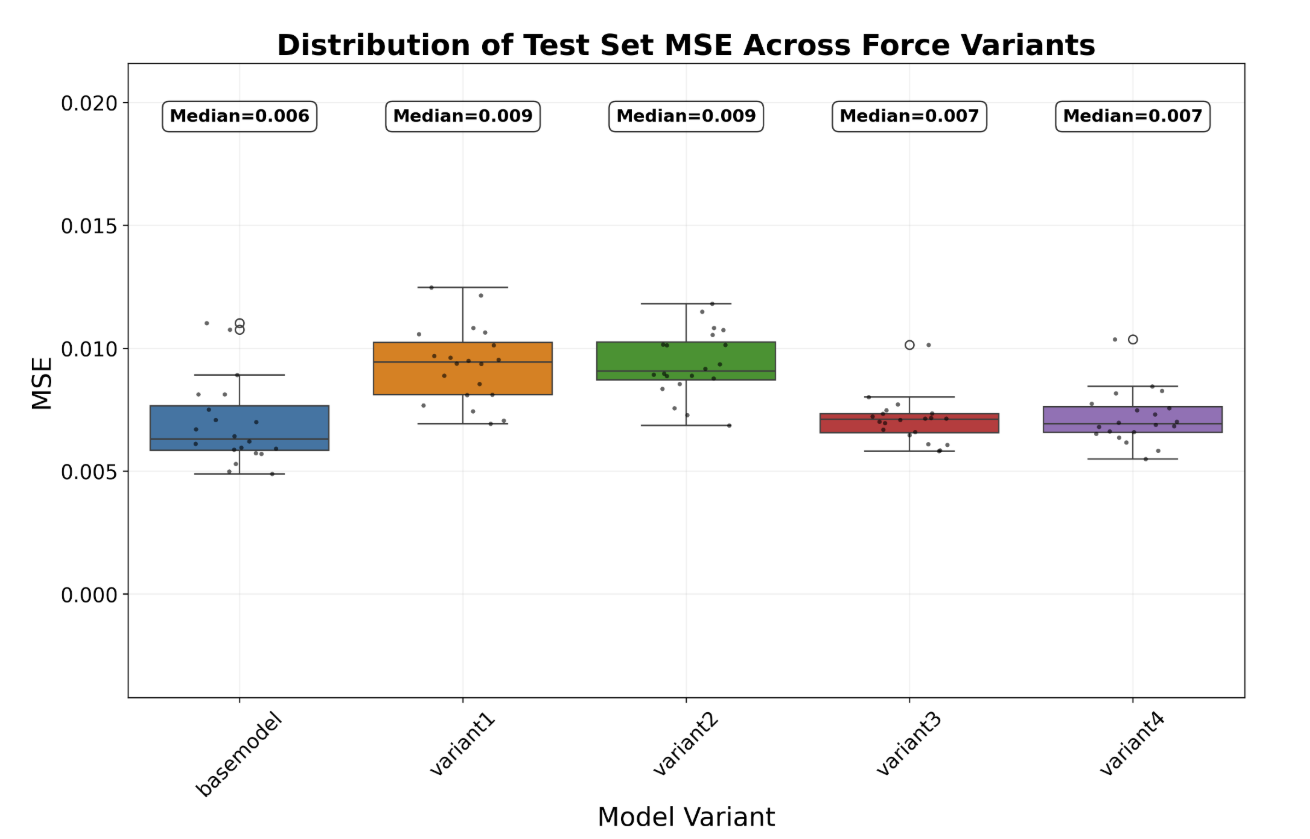}
    \vspace{-0.6cm}
    \caption{
    Force prediction test-set MSE over 20 training seeds for the expert baseline and four TacEvo-discovered variants; boxes show the seed-wise distribution and markers denote individual runs.
    }
\label{fig:force_boxplot}
\end{figure}

Fig.~\ref{fig:force_boxplot} shows the distribution of test-set MSE values over 20 independent training seeds for the baseline and the four TacEvo variants on Force Prediction, where lower is better. 
In Figs.~\ref{fig:force_boxplot} and \ref{fig:grating_boxplot}, overlaid markers denote individual seed runs; their horizontal offsets are jittered within each model category for visual clarity only and do not encode any additional variable.
The baseline achieves the lowest median MSE (0.006), but the key question is whether differences are statistically meaningful across seeds. 
Table~\ref{tab:force_mwu} summarises Mann-Whitney U tests with Holm-Bonferroni correction. 
Variants~1 and~2 are differ from the expert baseline ($p_{\mathrm{adj}}{<}0.001$), whereas Variants~3 and~4 are statistically indistinguishable from the baseline ($p_{\mathrm{adj}}{=}0.25$ and $0.40$). 
Combined with the MSE distributions in Fig.~\ref{fig:force_boxplot}, this indicates that TacEvo discovers multiple architecturally distinct solutions, two of which remain competitive with the expert-designed model on force prediction.

\subsubsection{Post-search High-fidelity Evaluation}
\label{subsec:grating_results}

\begin{figure}[!t]
\centering
\includegraphics[width=1\linewidth]{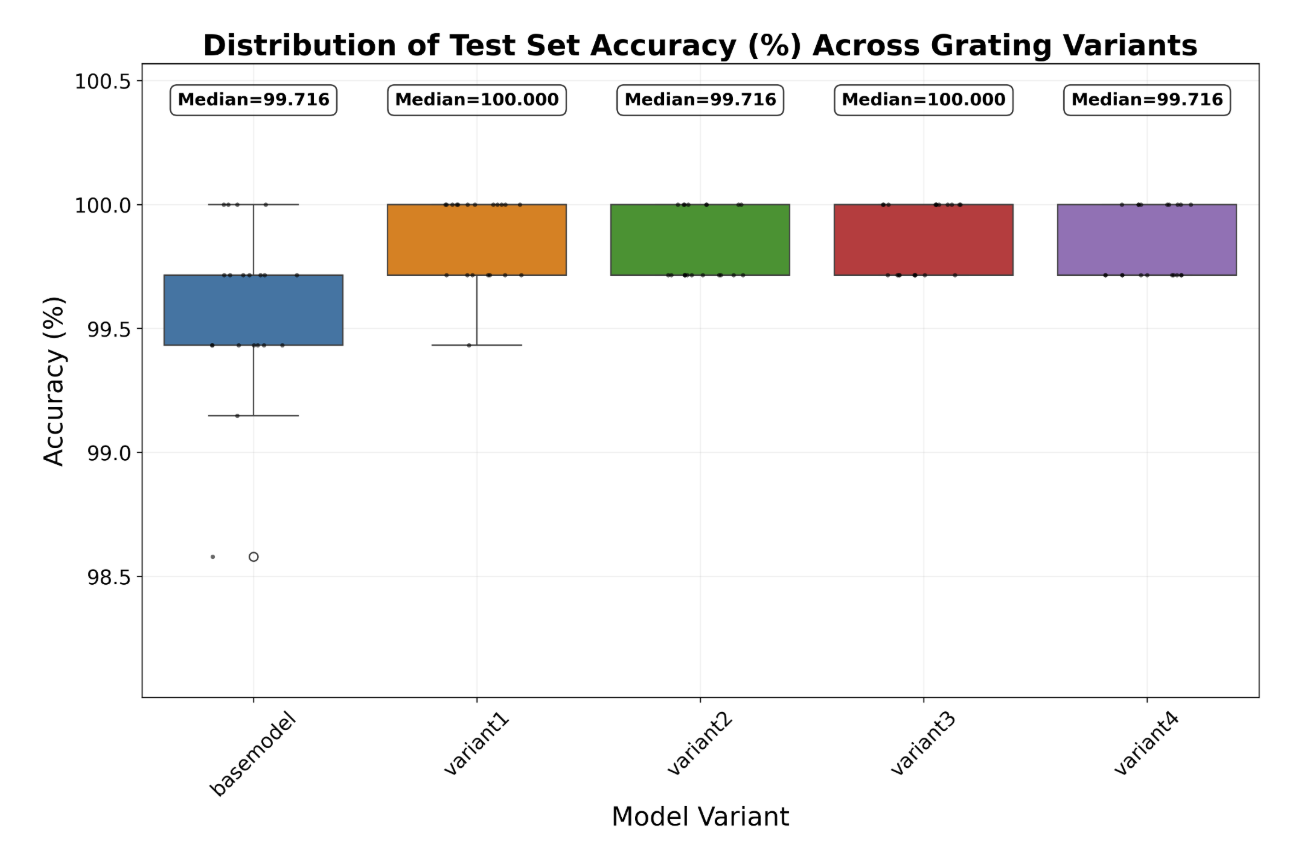}
\vspace{-0.6cm}
\caption{Grating classification test accuracy over 20 training seeds for the expert baseline and four TacEvo-discovered variants; boxplots show the seed-wise distribution with individual runs overlaid.}
    \vspace{-0.4cm}
\label{fig:grating_boxplot}
\end{figure}

Fig.~\ref{fig:grating_boxplot} reports test-set accuracy distributions over 20 seeds for Grating Classification, where higher is better. All four TacEvo variants achieve higher median accuracy than the baseline; Variants~1 and~3 reach a perfect median accuracy of 100.0\%, while the baseline median is 99.7\%. Importantly, these improvements remain statistically significant under non-parametric testing with Holm--Bonferroni correction (Table~\ref{tab:grating_mwu}, all $p_{\mathrm{adj}}{<}0.05$). 
This proves that TacEvo can autonomously discover architectures that outperform a human-designed baseline on a fine-grained tactile discrimination task.


\begin{figure*}[!t]
    \centering
\includegraphics[width=1\linewidth]{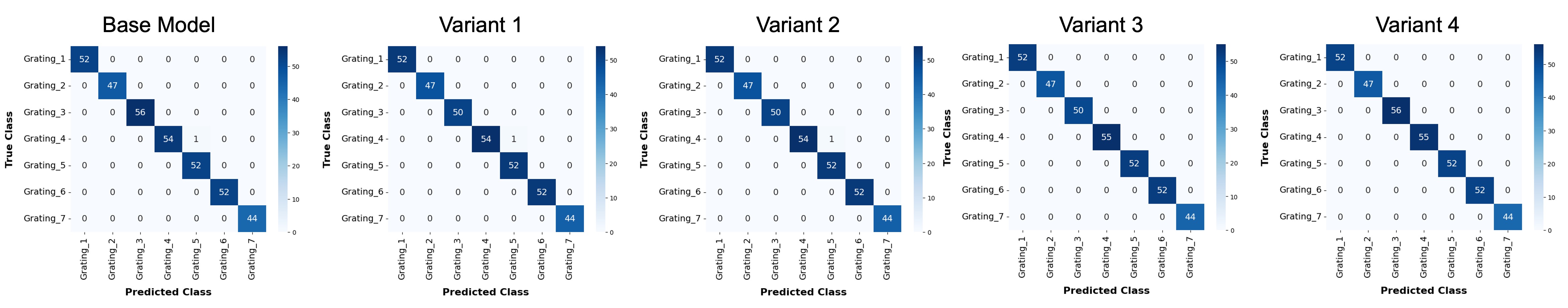}
\vspace{-0.7cm}
\caption{Test-set confusion matrices for grating classification for the baseline and four TacEvo variants on the same seed, showing per-class performance across the seven grating classes.}
    \vspace{-0.4cm}
\label{fig:grating_confusion}
\end{figure*}

To inspect per-class behaviour, Fig.~\ref{fig:grating_confusion} shows confusion matrices on the test set for a representative seed. All models present strong diagonals, indicating robust generalisation. 
In this seed, the baseline and Variants~1-2 exhibit a single misclassification, whereas Variants~3-4 achieve accurate classification, further supporting the high accuracy observed in the seed-aggregated evaluation.

\subsubsection{Qualitative Summary of Discovered Architectural Motifs}
\label{subsec:motifs}
As shown in Table~\ref{tab:motifs}, for force prediction, competitive variants frequently introduce lightweight spatial attention and replace max pooling with strided convolutions, alongside modest changes in convolutional depth. For Grating Classification, the most successful variants tend to widen convolutional layers, deepen the classifier head, replace ReLU with GELU, add residual connections, and remove dropout while adding normalisation in the head, which suggests that fine-grained tactile texture discrimination benefits from higher-capacity and better-conditioned feature processing. 

\begin{table}[!t]
\centering
\caption{
Mann–Whitney U tests on Force Prediction test MSE comparing each TacEvo variant against the expert baseline across 20 seeds.}
\label{tab:force_mwu}
\footnotesize
\setlength{\tabcolsep}{4pt}
\begin{tabular}{lcc c}
\hline
Comparison & $p$ & $p_{\mathrm{adj}}$ & Sig. ($\alpha{=}0.05$) \\
\hline
Variant 1 vs Baseline & $<0.001$ & $<0.001$ & Yes \\
Variant 2 vs Baseline & $<0.001$ & $<0.001$ & Yes \\
Variant 3 vs Baseline & $0.25$ & $0.25$ & No \\
Variant 4 vs Baseline & $0.20$ & $0.40$ & No \\
\hline
\end{tabular}
\end{table}

\begin{table}[!t]
\centering
\caption{Mann-Whitney U test for grating classification, comparing each variant against the baseline. }
\label{tab:grating_mwu}
\footnotesize
\setlength{\tabcolsep}{4pt}
\begin{tabular}{lcc c}
\hline
Comparison & $p$ & $p_{\mathrm{adj}}$ & Sig. ($\alpha{=}0.05$) \\
\hline
Variant 1 vs Baseline & $0.0018$ & $0.0055$ & Yes \\
Variant 2 vs Baseline & $0.0064$ & $0.0064$ & Yes \\
Variant 3 vs Baseline & $0.0015$ & $0.0060$ & Yes \\
Variant 4 vs Baseline & $0.0041$ & $0.0082$ & Yes \\
\hline
\end{tabular}
\vspace{-0.2cm}
\end{table}

\begin{table}[!t]
\centering
\caption{Motif-level summary of recurrent architectural changes observed among the top TacEvo-discovered variants for force prediction and grating classification.}
\label{tab:motifs}
\footnotesize
\begin{tabular}{p{0.22\columnwidth} p{0.72\columnwidth}}
\hline
Task & Recurrent motifs in top variants \\
\hline
Force regression & Spatial attention; maxpool $\rightarrow$ strided conv; depth/shape tweaks (2--4 conv blocks); occasional $1{\times}1$ conv refinements. \\
Grating classification & Wider conv layers; deeper FC head; ReLU $\rightarrow$ GELU; residual blocks; dropout removed; BN/LN added in head. \\
\hline
\end{tabular}
\vspace{-0.4cm}
\end{table}

\section{Conclusion}
We introduced TacEvo, a self-evolving architecture discovery framework for vision-based tactile sensing that closes the loop between code-level hypothesis generation and downstream task feedback. 
By combining an LLM as a mutation/crossover operator with quality-diversity search and a dual-archive memory (architectures and prompting strategies), TacEvo autonomously produces a diverse repertoire of high-quality tactile models rather than converging to a single design. 
Across two tasks, TacEvo reliably generates trainable networks, improves search fitness over generations, and discovers architectures that match the expert baseline on force prediction while outperforming it on fine-grained grating classification. 
This work shows that LLM-driven, feedback-grounded quality-diversity search is a practical AI-for-scientific-discovery mechanism for specialised robotic sensing, enabling automated, continual refinement of both architectures and the generation policy that proposes them.

\bibliographystyle{IEEEtran}
\bibliography{references}

\end{document}